\documentclass[conference,compsoc]{IEEEtran}
\IEEEoverridecommandlockouts

\usepackage{url}
\usepackage{cite}
\usepackage{amsmath,amssymb,amsfonts}
\usepackage{graphicx}
\usepackage{booktabs}
\usepackage{placeins}   
\usepackage{float}
\usepackage{textcomp}
\usepackage{xcolor}
\usepackage[hidelinks]{hyperref}

\begin{document}

\title{A Three-Axis Stress Test
of LLM vs Classical ML for Network Intrusion Detection under
Distribution Shift and Adversarial Evasion}





\author{
\IEEEauthorblockN{Muhammad Ebad Atif}
\IEEEauthorblockA{\textit{Dhanani School of Science and Engineering} \\
\textit{Habib University}\\ Karachi, Pakistan \\ ma09639@st.habib.edu.pk}
\and
\IEEEauthorblockN{Muhammad Haider Ali}
\IEEEauthorblockA{\textit{Faculty of Computer Sciences and Engineering} \\
\textit{Ghulam Ishaq Khan Institute of Engineering Sciences}\\ Swabi , Pakistan \\ u2024385@giki.edu.pk}
}

\maketitle

\footnotetext[0]{Code and reproduction materials:
\url{https://github.com/ebadatif/IDS-three-axis-evaluation}}

\begin{abstract}
Large language models are increasingly benchmarked against classical
machine learning for network intrusion detection (NIDS), almost
always using same-dataset evaluation, and that protocol turns out to
be incomplete. Evaluating XGBoost and RoBERTa-LoRA on two
independently collected NetFlow v2 networks across three axes
(same-dataset performance, cross-dataset transfer, and adversarial
evasion) reveals no universal winner. The two models are statistically
tied same-dataset. XGBoost wins decisively under cross-dataset
distribution shift, by 15 points of F1 and 25 points of balanced
accuracy; on the target network RoBERTa-LoRA's false positive rate
reaches 0.78, leaving it barely above chance despite a superficially
moderate F1. RoBERTa-LoRA wins decisively under adversarial evasion,
by roughly 17 points of F1 at a representative mid-range perturbation
strength, while both models hold false positive rates below 0.01
throughout. The model an evaluator would recommend therefore depends
entirely on which axis is tested, not on same-dataset accuracy alone.
A staged feature-leakage ablation improves cross-dataset transfer
non-monotonically, indicating the leakage signal is distributed
across the feature representation rather than confined to a few
columns, and cross-dataset transfer between our two networks is
strongly directional. These results argue for evaluating NIDS models
along multiple independent robustness axes, and with more than one
metric per axis.
\end{abstract}


\section{Introduction}
\label{sec:introduction}

Machine learning-based Network Intrusion Detection Systems (NIDSs)
are increasingly benchmarked against large language models, typically
using a same-dataset protocol: train and test on splits from a single
network. This is convenient, but it says little about two conditions
that matter in practice: deployment on a different network and an
adversary deliberately crafting evasive traffic. Same-dataset
evaluation exercises neither. The risk of overstating what such an
evaluation demonstrates for real-world NIDS deployment is a
long-standing concern, dating back to Sommer and
Paxson~\cite{sommer2010closedworld}'s critique of closed-world
evaluation in ML-based intrusion detection. Recent surveys catalogue
the temporal leakage, weak splitting protocols, and poor
cross-dataset generalization that inflate reported NIDS
results~\cite{elmahdaouy2026}.

Recent work moves past same-dataset evaluation, but only one axis at
a time. Mehavilla et al.~\cite{mehavilla2026} compare LLMs and
classical models same-dataset, finding XGBoost wins accuracy while
LLMs are more data-efficient. Bui/Boffa et al.~\cite{bui2024conext}
test generalization within a single network via a time-based,
zero-day split, finding fine-tuned BERT beats classical ML
in-distribution but degrades sharply under attack drift. Neither
evaluates cross-dataset transfer \emph{between independently
collected networks}, nor adversarial evasion. Bui/Boffa themselves
flag zero-day generalization within their own dataset as the most
urgent open direction, leaving the two axes we examine outside their
reported scope. This gap matters because, as we show, these two axes
can yield opposite conclusions about which model is preferable; a
verdict from either alone risks being incomplete.

This paper evaluates XGBoost and RoBERTa-LoRA across both axes, on
two independently collected NetFlow v2 networks, under one shared
harness. No model wins universally: the two are statistically tied
same-dataset, XGBoost wins decisively under cross-dataset shift, and
RoBERTa-LoRA wins decisively under adversarial evasion by roughly 17
F1 points at a representative mid-range perturbation strength. Which
model an evaluator would recommend therefore depends on which axis
they choose to test. A secondary methodological finding runs through
both axes: F1 alone is insufficient to characterize either result,
since on the cross-dataset axis it conceals a near-chance classifier
behind a moderate score, while on the adversarial axis only the false
positive rate distinguishes robustness from indiscriminate flagging.
Two additional findings arose as the work progressed. A staged
feature-ablation study shows that leakage removal does not improve
cross-dataset transfer monotonically, indicating the leakage signal
is distributed rather than confined to a few columns. Cross-dataset
transfer between our two networks is also strongly directional, a
pattern related to differing attack character across networks,
though this mechanism was not directly verified
(Sec.~\ref{sec:discussion}).

Sec.~\ref{sec:related-work} positions this work against prior
LLM-NIDS studies. Sec.~\ref{sec:datasets}--\ref{sec:eval_prot} cover
datasets, ablation, models, and evaluation protocol.
Sec.~\ref{sec:results-summary} reports results across all three
axes. Sec.~\ref{sec:discussion} argues the paper's thesis and
explains the ablation and transfer findings.
Sec.~\ref{sec:limitations} states scope constraints, and
Sec.~\ref{sec:conclusion} closes the paper.

\section{Related Work}
\label{sec:related-work}

Whether large language models are competitive with classical machine
learning for network intrusion detection is a growing question, but
existing studies each stress-test only one axis of robustness at a
time. This section positions the present work against the two most
directly comparable studies.

\paragraph{Same-dataset comparison: Mehavilla et al.}
Mehavilla et al.~\cite{mehavilla2026} evaluate LLMs directly as flow
classifiers, comparing small decoder-only transformers (GPT-2,
GPT-Neo-125M, LLaMA-3.2-1B) against classical machine learning
(Random Forest, XGBoost) and deep learning baselines (MLP, GRU,
LeNet-5) on the CIC-IoT-2023 dataset. Their evaluation is entirely
same-dataset: models are trained and tested on splits drawn from the
same underlying traffic distribution. XGBoost wins on accuracy
(F1 $\approx$ 0.97 on their multiclass task), narrowly ahead of the
best fine-tuned LLM ($\approx$ 0.96). Their one clear LLM advantage
is data efficiency: pre-trained LLMs are already near their
performance ceiling with 10,000 samples per class, while the deep
learning baselines need substantially more data to reach comparable
performance. Their study does not evaluate cross-dataset transfer or
adversarial robustness -- a limitation the authors themselves note,
alongside their use of only three small LLMs and single-run
(non-multi-seed) results.

\paragraph{Distribution shift: Bui/Boffa et al.}
Bui/Boffa et al.~(CoNEXT 2024)~\cite{bui2024conext} take a
complementary angle, comparing fine-tuned BERT against a large
classical ML sweep (over 50 model/hyperparameter combinations) under
both a standard stratified split and a time-based, zero-day split
(their 15/08-split) designed to test generalization to novel attacks.
In-distribution, their fine-tuned BERT consistently beats the best
classical model, by a mean pairwise difference of 5 percentage points
of weighted accuracy (95\% CI [0.04, 0.06]). Under their zero-day
split, performance degrades sharply for every model. BERT falls from
98\% to 79\% weighted accuracy in their most information-rich
IDS-assistance configuration (payload + 5-tuple + eventName inputs),
and the drop is comparable in less-favourable configurations
(e.g.\ 88\% to 73\% for the IDS-replacement setting with payload +
5-tuple). They further find that scaling to a larger LLM (Mistral-7B)
provides no measurable benefit over the much smaller BERT-base
($-0.0\%$ relative to the BERT baseline in their ablation).
Domain-specific pretraining (SecureBERT, UniXcoder) yields only
marginal gains ($+0.7\%$ and $+1.1\%$ respectively), and they
conclude that a small, well-fine-tuned model is the practical sweet
spot. Like Mehavilla et al., their study does not include an
adversarial evasion axis, and the authors flag zero-day
generalization within their dataset as the most urgent open
direction, not cross-dataset transfer between independently collected
networks.

Between these two studies, three evaluation axes for LLM-vs-classical
NIDS comparison have each been examined once: same-dataset accuracy
(Mehavilla et al.), zero-day/temporal generalization within a single
network (Bui/Boffa et al.), and, separately, general-purpose
prompting and RAG-based approaches with frozen LLMs (also evaluated
by Bui/Boffa et al., and found to substantially underperform
fine-tuning). Neither study, nor, to our knowledge, any other work in
this space, evaluates cross-dataset transfer between two
independently collected networks \emph{and} adversarial evasion
robustness on the same pair of matched models. This gap matters
because these two axes, as Sec.~\ref{sec:discussion} shows, can
produce opposite rankings between the same two models: a comparison
run on only one of them risks generalizing a conclusion (``XGBoost
beats RoBERTa'' or ``RoBERTa beats XGBoost'') that does not hold under
the other. The present study addresses this by evaluating XGBoost and
RoBERTa-LoRA under a single, shared harness across same-dataset,
bidirectional cross-dataset transfer, and adversarial evasion, on the
same NetFlow v2 feature schema throughout.

The cross-dataset finding reported here (Sec.~\ref{sec:results-cd}),
near-perfect same-dataset performance that collapses under
distribution shift, is not an isolated result. Cantone et
al.~\cite{cantone2024crossdataset} independently document the same
pattern using four classifiers across four NIDS datasets, and their
work is, to our knowledge, the closest existing confirmation of the
effect measured here between NF-UNSW-NB15-v2 and
NF-CSE-CIC-IDS2018-v2. Related work reports similar generalization
failure in adjacent settings, including enterprise
datasets~\cite{iiot_crossdomain_2024}, IIoT network deployments, and
unsupervised NIDS models. Several of these call for standardized
cross-evaluation frameworks and explainability-driven analysis of
what fails to transfer and why. Separately, a recent comparison of
tabular representation learning against transformer baselines for
NIDS finds no single approach consistently dominates across
evaluation scenarios~\cite{tabular_repr_2605}, a parallel to the
``no universal winner'' finding of Sec.~\ref{sec:discussion} obtained
along a different pair of evaluation axes.

\section{Methodology}

The experimental design follows the methodological pitfalls
catalogued by Arp et al.~\cite{arp2022dosdonts} for machine learning
in computer security: avoiding data snooping between train and test
splits (Sec.~\ref{sec:preprocessing}), auditing for spurious,
non-causal features before reporting results
(Sec.~\ref{sec:feature_abl_study}), and evaluating with a realistic,
deployment-relevant protocol rather than same-dataset accuracy alone
(Sec.~\ref{sec:eval_prot}).

\subsection{Datasets}
\label{sec:datasets}

We use two datasets from the University of Queensland NetFlow v2
collection: NF-UNSW-NB15-v2 and NF-CSE-CIC-IDS2018-v2
\cite{sarhan2022standard}. Both are re-extracted from the original
raw PCAP captures of UNSW-NB15 and CSE-CIC-IDS2018 into an identical,
standardized schema of 43 NetFlow features, in the same column order.
That shared schema makes a fair cross-dataset comparison possible.
Without it, any transfer result would be confounded by differing
feature definitions rather than reflecting a distribution shift
between networks.

NF-UNSW-NB15-v2 contains approximately 2.39M flows, of which 3.98\%
are attack traffic. NF-CSE-CIC-IDS2018-v2 is substantially larger, at
approximately 18.9M flows, with 11.95\% attack traffic, roughly a
threefold difference in attack prevalence (addressed in
Sec.~\ref{sec:preprocessing} via balanced sampling).

The two datasets originate from different institutions, years, and
network configurations. UNSW-NB15 was generated at the Australian
Centre for Cyber Security (UNSW Canberra), while CSE-CIC-IDS2018 was
generated by the Canadian Institute for Cybersecurity. That
institutional and temporal separation makes the pair a valid basis
for testing cross-dataset generalization, as opposed to evaluating on
a held-out split of the same collection process.

\paragraph{Attack taxonomy mismatch.}
The two datasets' attack categories are almost entirely disjoint.
UNSW-NB15 attacks are labeled as Exploits, Fuzzers, Reconnaissance,
Generic, DoS, Shellcode, Backdoor, Analysis, and Worms.
CSE-CIC-IDS2018 attacks are labeled as DDoS variants (HOIC, LOIC-HTTP,
LOIC-UDP), DoS variants (Hulk, GoldenEye, Slowloris, SlowHTTPTest),
Infiltration, Bot, Brute Force (SSH, FTP, Web, XSS), and SQL
Injection. Only the coarse category ``DoS'' overlaps even loosely
between the two label sets. Because the label spaces are effectively
disjoint, multiclass cross-dataset evaluation is not meaningful: a
model cannot be evaluated on attack subtypes it never saw during
training. The task is therefore collapsed to binary classification
(benign vs.\ attack) for all cross-dataset experiments, the only
label granularity shared by both datasets. The pipeline consequently
records only the aggregate binary confusion matrix in every
condition, which constrains the mechanistic analysis in
Sec.~\ref{sec:discussion}.

Before any modeling was performed, identically named columns in the
two datasets turned out to be stored with different data types.
\texttt{IN\_PKTS} is stored as \texttt{int16} in NF-UNSW-NB15-v2 but
\texttt{int32} in NF-CSE-CIC-IDS2018-v2, and
\texttt{SRC\_TO\_DST\_SECOND\_BYTES} is \texttt{float32} versus
\texttt{float64}. Since each dataset's original storage type was
chosen to fit its own value range, this divergence is itself evidence
of a real distributional difference: CSE-CIC-IDS2018 contains flows
with packet and byte counts that exceed the range representable by
UNSW-NB15's narrower types. This is treated as an independent,
pre-modeling signal that the cross-dataset shift measured later on is
real and not an artifact of the pipeline.

\subsection{Preprocessing}
\label{sec:preprocessing}

\paragraph{Balanced sampling.}
Attack prevalence differs roughly threefold between the two datasets
(Sec.~\ref{sec:datasets}). Because precision and recall are both
sensitive to base rate, evaluating models across datasets with
different prevalence would confound two distinct effects: a shift in
the underlying traffic distribution, and a shift in how often the
positive class occurs. To isolate the former, balanced subsets are
drawn from both datasets (50,000 flows per class) so that attack
prevalence is identical in every training and test split. Any drop in
cross-dataset performance reported in Sec.~\ref{sec:results-cd} can
therefore be attributed to distribution shift, not to a difference in
class balance between source and target. This mirrors the
balanced-sampling approach used by Mehavilla et
al.\ \cite{mehavilla2026}. The balanced protocol places attack
prevalence far above any realistic operational base rate, so absolute
precision figures throughout this paper should be read as comparative
between models rather than as deployment estimates. This is why false
positive rate is reported alongside precision.

The dtype divergence noted above is informative as evidence of
distribution shift, but it is also a practical obstacle: models
cannot be trained on features that are not uniformly typed across
datasets. Non-finite and extreme values are capped at the 99.9th
percentile of each feature, computed from the UNSW (training) side
only to avoid test-set leakage, and the same caps are applied to CIC.
Every feature is then cast to \texttt{float32} at load time, so that
both datasets present an identical, bounded numeric schema to every
downstream model.

Before any leakage-driven feature ablation
(Sec.~\ref{sec:feature_abl_study}), four columns are dropped outright:
\texttt{L4\_SRC\_PORT}, \texttt{L4\_DST\_PORT},
\texttt{DNS\_QUERY\_ID}, and the multiclass \texttt{Attack} label
(retained separately for post-hoc analysis only, never as a feature).
The three non-label columns are session identifiers. A source or
destination port number, or a DNS query ID, describes which specific
connection a flow belongs to, saying nothing about what that
connection did. Retaining them would risk the model memorizing
connection identity instead of learning transferable attack
behavior, the failure mode the rest of the feature analysis is
designed to guard against.

The task is framed as binary classification. The \texttt{Label}
column (benign vs.\ attack) is the prediction target; the
\texttt{Attack} column, which encodes the fine-grained multiclass
attack subtype, is dropped from the feature set entirely. As
established in Sec.~\ref{sec:datasets}, this binary framing is a
requirement of the disjoint label spaces, so no multiclass target
could be shared between the two datasets.

Random seeds (42--45) are built into the data loading and model
training pipeline from the start. Every classical model result in
this paper is obtained by re-running the full pipeline, sampling
through training to evaluation, under each of the four seeds
independently. This was a deliberate design decision: multi-seed
evaluation is straightforward to build in from the outset and
difficult to retrofit cleanly, particularly once balanced sampling
introduces its own source of randomness.

\subsection{Feature Ablation Study}
\label{sec:feature_abl_study}

Time-to-live (TTL) is a counter set by the sending operating system
at a fixed default (commonly 64 for Linux/Mac, 128 for Windows, or
255 for some network hardware) and decremented by one at each network
hop. The value observed at the receiving end therefore encodes which
operating system sent the packet and how many hops away it
originated -- information about the network's physical and software
topology, and nothing about the content or intent of the traffic. In
NF-UNSW-NB15-v2, benign and attack traffic were generated by
different machines during dataset construction. A model can
consequently learn a rule of the form ``TTL = X implies benign, TTL =
Y implies attack'' that is entirely accurate on this dataset while
reflecting nothing about attack behavior: it has memorized which
machine generated a flow. This single, concrete leakage mechanism
motivated a systematic search for other features exhibiting the same
problem.

To flag ablation candidates, each feature's solo predictive power
against the label is computed using a one-level decision stump (a
single-feature threshold split) fit on that feature alone and scored
on held-out data. A high score marks a feature for investigation; solo
F1 $> 0.90$ is used as the flag threshold, and a flag triggers
scrutiny rather than automatic removal. The audit was run iteratively,
re-computed after each round of removals, to check whether remaining
features still stood in for the label on their own. Two other
diagnostics accompanied the solo-F1 stump: full-model F1 on the
candidate feature set, to catch collapse before it reached
cross-dataset testing, and XGBoost's built-in feature-importance
ranking, to check whether the model leaned on one or two features
disproportionately.

Applying this audit to the full 38-feature set surfaced three groups
of suspects, grouped by the source and pattern of their solo
predictive power. The first is TTL (\texttt{MIN\_TTL},
\texttt{MAX\_TTL}), flagged via the mechanism described above. The
second is the packet-length features (\texttt{MIN\_IP\_PKT\_LEN},
\texttt{MAX\_IP\_PKT\_LEN}, \texttt{SHORTEST\_FLOW\_PKT},
\texttt{LONGEST\_FLOW\_PKT}), fixed per-flow size statistics that can
reflect equipment or capture configuration as easily as behavior. The
third is a smaller group of remaining high-solo-F1 features
(\texttt{SERVER\_TCP\_FLAGS}, \texttt{TCP\_WIN\_MAX\_IN},
\texttt{TCP\_WIN\_MAX\_OUT}). For none of these three groups could an
environmental fingerprint be cleanly separated from discriminative
behavior using solo F1 alone -- a feature can score highly for either
reason, and the stump test does not distinguish them. The three
groups were therefore treated as three competing hypotheses about
where the leakage boundary sits, resolved empirically by testing each
group's removal against cross-dataset transfer
(Sec.~\ref{sec:results-cd}). Cross-dataset F1, not same-dataset solo
F1, is the arbiter of whether a suspected feature was leaking.

Corresponding to the three suspect groups above, three progressively
larger removals were evaluated, each scored on both same-dataset and
cross-dataset performance:
\begin{itemize}
    \item \textbf{Set A} (36 features): drop TTL only.
    \item \textbf{Set B} (32 features): drop TTL and packet-length features.
    \item \textbf{Set C} (29 features): an aggressive drop, additionally removing \texttt{SERVER\_TCP\_FLAGS}, \texttt{TCP\_WIN\_MAX\_IN}, and \texttt{TCP\_WIN\_MAX\_OUT}.
\end{itemize}
Staging the removals shows how transfer performance responds at each
step. All three sets are evaluated under an identical protocol, train
on the full source dataset and test on the full target dataset,
matching the cross-dataset harness of Sec.~\ref{sec:eval_prot}, so
that ablation results and benchmark results are directly comparable.

Table~\ref{tab:ablation} shows that the ablation did not improve
cross-dataset transfer monotonically. Set A left cross-dataset F1 low
(0.0488, FNR 0.9656). Set B made transfer worse, dropping
cross-dataset F1 to 0.0282 with a false-negative rate of 0.9851,
despite removing more candidate leakage features than Set A. Only Set
C recovered cross-dataset performance (F1 0.8116, FNR 0.1293).
Throughout all three sets, same-dataset F1 barely moved: 0.9972 in
Sets A and B, 0.9966 in Set C, a difference of six ten-thousandths.
Because same-dataset performance is insensitive to which leakage
features are present, the leakage signal cannot be isolated to one or
two columns -- it is distributed across the feature representation.
Removing part of it (Set B) can disturb a competing, transferable
signal while leaving enough leakage-adjacent structure intact to
mislead the model on the target domain.

Set C (29 features) is locked as the final feature schema. Every
cross-dataset and adversarial-evasion result reported in this paper
(Sec.~\ref{sec:results-cd}, Sec.~\ref{sec:results-adv}) uses Set C
exclusively; same-dataset results in Table~\ref{tab:same-dataset} are
also reported under Set C for consistency across all conditions.

\subsection{Models}
\label{sec:models}

The primary classical model is XGBoost, trained with 100 trees,
\texttt{n\_jobs=-1}, and \texttt{eval\_metric=logloss}. Every XGBoost
result reported in this paper is obtained across four seeds (42--45),
using the locked Set C feature schema. Random Forest is used as a
secondary classical baseline, trained under the same seed set and
feature schema, to check whether findings attributed to XGBoost are
specific to gradient boosting or hold across tree-based methods
generally.

The LLM model is RoBERTa-base, fine-tuned with LoRA adapters, using
rank $r=8$, scaling factor $\alpha=16$, with adapters applied to the
query, key, and value projection matrices, leaving approximately
0.82\% of total model parameters trainable. Each of the two LLM arms
(UNSW-trained and CIC-trained) is fine-tuned in a single run from a
freshly loaded pretrained checkpoint with a freshly initialised
adapter, so neither arm carries optimization history from the other
or from any earlier run. This is stated explicitly because prior
work~\cite{mehavilla2026} found that fine-tuning from pretrained
weights versus training from scratch materially affects downstream
performance.

RoBERTa operates on text, so each flow record is serialized into a
string prior to tokenization. Every feature in the Set C schema is
rendered as a \texttt{FEATURE=value} pair, and these pairs are
concatenated into a single string representing one flow. This step is
specific to the LLM arm; XGBoost and Random Forest consume the same
Set C features directly as numeric vectors, with no text conversion.

Serializing feature names into the input text gives RoBERTa an
explicit structural signal, the identity of each feature, that the
numeric-only classical models do not receive. XGBoost and Random
Forest see only feature values, positioned by column order, with no
access to feature names at inference time. The two arms therefore do
not operate on strictly identical information, and any performance
comparison between them
(Sec.~\ref{sec:results-sd}--\ref{sec:results-adv}) should be read
with this asymmetry in mind. Sec.~\ref{sec:discussion} returns to it
as the leading candidate explanation for RoBERTa-LoRA's comparative
stability under adversarial evasion.

Unlike the four-seed classical arm, RoBERTa is trained with a single
seed per condition, due to the compute cost of LoRA fine-tuning
across every condition in the study. RoBERTa results are therefore
reported as point estimates rather than mean $\pm$ std throughout;
Sec.~\ref{sec:limitations} revisits the implications.

\subsection{Adversarial Evasion Attack}
\label{sec:adv_evas_attack}

Robustness to adversarial evasion is evaluated using a feature-space
``drift toward benign'' attack. Let $\mathbf{x}$ denote the
perturbable-feature vector of an attack flow (the perturbable subset
is defined below), and let $\boldsymbol{\mu}$ denote the centroid of
benign flows in that same feature space,
\begin{equation}
\boldsymbol{\mu} = \frac{1}{|\mathcal{B}|} \sum_{\mathbf{b} \in \mathcal{B}} \mathbf{b},
\label{eq:centroid}
\end{equation}
computed over the benign flows $\mathcal{B}$ in the training split
only, so that the attacker is not given access to held-out data. For
a perturbation strength $\varepsilon \in [0,1]$, the perturbed flow is
generated by linear interpolation toward this centroid,
\begin{equation}
\mathbf{x}'(\varepsilon) = (1-\varepsilon)\,\mathbf{x} + \varepsilon\,\boldsymbol{\mu},
\label{eq:interp}
\end{equation}
followed by projection through the domain-constraint operator
$\Pi(\cdot)$ described below, giving the final perturbed flow
$\Pi(\mathbf{x}'(\varepsilon))$. Ten perturbation strengths are swept,
where $\varepsilon=0$ reproduces the original attack flow
($\mathbf{x}'(0) = \mathbf{x}$) and $\varepsilon=1$ moves the flow
fully to the benign centroid ($\mathbf{x}'(1) = \boldsymbol{\mu}$).
Constrained feature-space perturbation of this kind is an established
methodology for approximating realistic evasion without requiring a
full problem-space attack
implementation~\cite{explainable_transferable_2401}, and the same
rationale is adopted here.

Naive linear interpolation over raw feature values can produce flows
that are not physically realizable, such as negative byte counts or
fractional packet counts. To keep perturbed flows within a plausible
feature space, the projection operator $\Pi(\cdot)$ in
Eq.~\eqref{eq:interp} enforces four constraints: non-negativity of
all features, integer-valued packet and byte counts, retransmitted-byte
counts bounded above by total byte counts, and fixed categorical
features (protocol, L7 protocol, ICMP type, DNS query type, FTP
return code) left unperturbed, i.e.\ $\Pi(\mathbf{x}')_j = x_j$ for
those coordinates regardless of $\varepsilon$. Of the 29 Set C
features, 23 are perturbable under this scheme (the coordinates of
$\mathbf{x}$ in Eq.~\eqref{eq:interp}) and 6 are held fixed. These
constraints do not guarantee that a perturbed flow is realizable by
an actual attacker in the underlying network protocol (see
Sec.~\ref{sec:limitations}), but they rule out perturbations that are
trivially infeasible.

At every $\varepsilon$, the evaluation set holds the 15{,}000
perturbed attack flows from the UNSW held-out split together with the
15{,}000 untouched benign flows from that same split. Only the
attacker's own traffic is perturbed; normal traffic is left
unmodified, as it would be in a real deployment. Including benign
traffic is essential: an attacks-only evaluation set makes precision
trivially unity by construction and renders false positives
unmeasurable. A model that simply labels everything an attack would
therefore appear maximally robust. F1, precision, recall and false
positive rate are therefore reported at every $\varepsilon$, with the
FPR column serving as the check on whether a high recall reflects
robustness or indiscriminate flagging.

By Eq.~\eqref{eq:interp}, $\mathbf{x}'(1) = \boldsymbol{\mu}$
regardless of the original flow $\mathbf{x}$: the interpolation
target is a single point, the benign centroid, so as
$\varepsilon \to 1$ every perturbed attack flow converges to the same
location in perturbable-feature space. This is quantified by counting
distinct perturbed attack rows at each $\varepsilon$: the
count is stable at 8{,}350 for every $\varepsilon \le 0.5$, falls to
8{,}349 at $\varepsilon = 0.7$, and collapses to \textbf{1} at
$\varepsilon = 1.0$. The $\varepsilon = 1.0$ condition therefore does
not measure robustness over 15{,}000 independent trials; it reports
one model decision replicated 15{,}000 times, with an effective
sample size of one. It is reported for completeness but excluded from
Fig.~\ref{fig:evasion-curve}, and no claim is based on it;
$\varepsilon \le 0.7$ is treated as the valid measurement range.

The adversarial evasion attack is evaluated same-dataset only, on
NF-UNSW-NB15-v2, using the locked Set C feature schema. Adversarial
evasion is not evaluated under cross-dataset transfer; the two stress
axes are examined independently in this study.

\subsection{Evaluation Protocol}
\label{sec:eval_prot}

Every model and every condition in this paper -- same-dataset,
cross-dataset in both directions, and adversarial evasion at all ten
$\varepsilon$ values -- is scored using one shared evaluation
function computing F1, precision, recall, balanced accuracy, false
positive rate (FPR), false negative rate (FNR), and throughput
(flows/sec). A single harness across the classical and LLM arms
guarantees that any performance difference reported in
Sec.~\ref{sec:results-sd}--\ref{sec:results-adv} reflects the models
themselves and not differences in scoring. As one consistency check,
the $\varepsilon=0$ row of the adversarial sweep reproduces the
same-dataset UNSW confusion matrix exactly for both models,
confirming that the two blocks evaluate the identical baseline.

The harness is applied under three conditions, which together define
the three axes of this study: a \emph{same-dataset} condition,
training on a 70\% split and testing on the held-out 30\% of the same
network; a \emph{cross-dataset} condition, training on the full
source network and testing on the full target network in both
directions; and an \emph{adversarial evasion} condition, testing
same-dataset performance against the perturbed attack flows described
in Sec.~\ref{sec:adv_evas_attack}. Sec.~\ref{sec:results-summary}
reports results for all three conditions and every model.

\section{Results}
\label{sec:results}

All results use the locked Set C feature schema (29 features) unless
otherwise noted. XGBoost and Random Forest are evaluated across four
seeds (42--45); RoBERTa-LoRA reflects a single training run per
condition due to compute cost (Sec.~\ref{sec:limitations}), so
RoBERTa figures are point estimates rather than mean~$\pm$~std.

\subsection{Same-Dataset Performance}
\label{sec:results-sd}

Table~\ref{tab:same-dataset} reports in-distribution performance:
train and test on the same network. Both models are effectively tied.
XGBoost reaches 0.9960--0.9971 F1 on UNSW-NB15 and 0.9764--0.9783 on
CIC-IDS2018 across the four seeds; RoBERTa-LoRA reaches 0.9938 and
0.9745 respectively. The gap never exceeds 0.004 F1 in either
direction.

\begin{table}[!tbp]
\caption{Same-dataset F1 (XGBoost: mean over 4 seeds, RoBERTa: single seed)}
\label{tab:same-dataset}
\centering
\footnotesize
\begin{tabular}{lccc}
\toprule
Dataset & XGBoost & RoBERTa-LoRA & Gap \\
\midrule
UNSW-NB15   & 0.9966 & 0.9938 & +0.0028 \\
CIC-IDS2018 & 0.9776 & 0.9745 & +0.0031 \\
\bottomrule
\end{tabular}
\end{table}

This is the number the field typically reports, and, as the following
two subsections show, it is also the least informative number for
predicting deployment behavior.

\begin{figure}[!tbp]
  \centering
  \includegraphics[width=\linewidth,height=0.4\textheight,keepaspectratio]{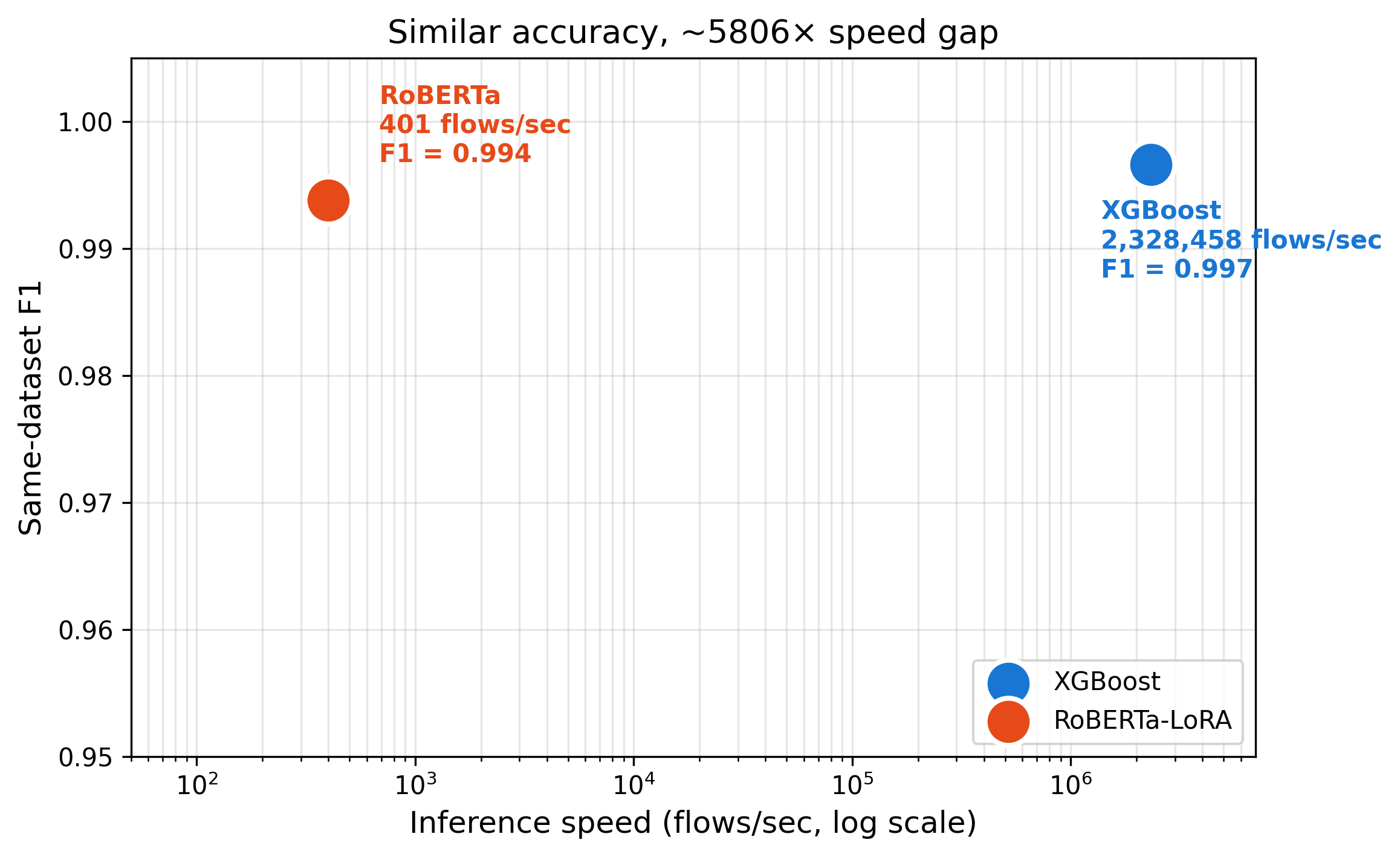}
  \caption{Near-identical same-dataset F1 at vastly different
  inference throughput (log scale). XGBoost processes roughly four
  orders of magnitude more flows per second than RoBERTa-LoRA.}
  \label{fig:speed}
\end{figure}

Beyond accuracy, the two arms differ sharply in inference throughput
(Fig.~\ref{fig:speed}). XGBoost processes on the order of $10^6$
flows/sec versus RoBERTa-LoRA's $\sim$$10^2$, an
$\sim$5{,}806$\times$ difference that bears directly on line-rate
deployment even though same-dataset accuracy is near-identical.
RoBERTa-LoRA throughput is measured end-to-end, including per-flow
text serialization and tokenization, both required in deployment;
XGBoost throughput reflects native numeric prediction on
already-vectorized input. The comparison is between complete
inference pipelines, not model forward passes alone.

\FloatBarrier

\subsection{Cross-Dataset Transfer}
\label{sec:results-cd}
\subsubsection{Feature ablation is non-monotonic}

Table~\ref{tab:ablation} shows the effect of progressively removing
candidate leakage features on XGBoost, evaluated both in-distribution
(SD) and under cross-dataset transfer (CD, UNSW$\to$CIC).

\begin{table}[!tbp]
\caption{Feature ablation: same-dataset vs.\ cross-dataset F1 (XGBoost, UNSW$\to$CIC)}
\label{tab:ablation}
\centering
\footnotesize
\begin{tabular}{lcccc}
\toprule
Feature set & \#Feat & SD F1 & CD F1 & CD FNR \\
\midrule
A: drop TTL only         & 36 & 0.9972 & 0.0488 & 0.9656 \\
B: drop TTL + pkt.\ len. & 32 & 0.9972 & 0.0282 & 0.9851 \\
C: aggressive drop       & 29 & 0.9966 & 0.8116 & 0.1293 \\
\bottomrule
\end{tabular}
\end{table}

\begin{figure}[!tbp]
  \centering
  \includegraphics[width=\linewidth]{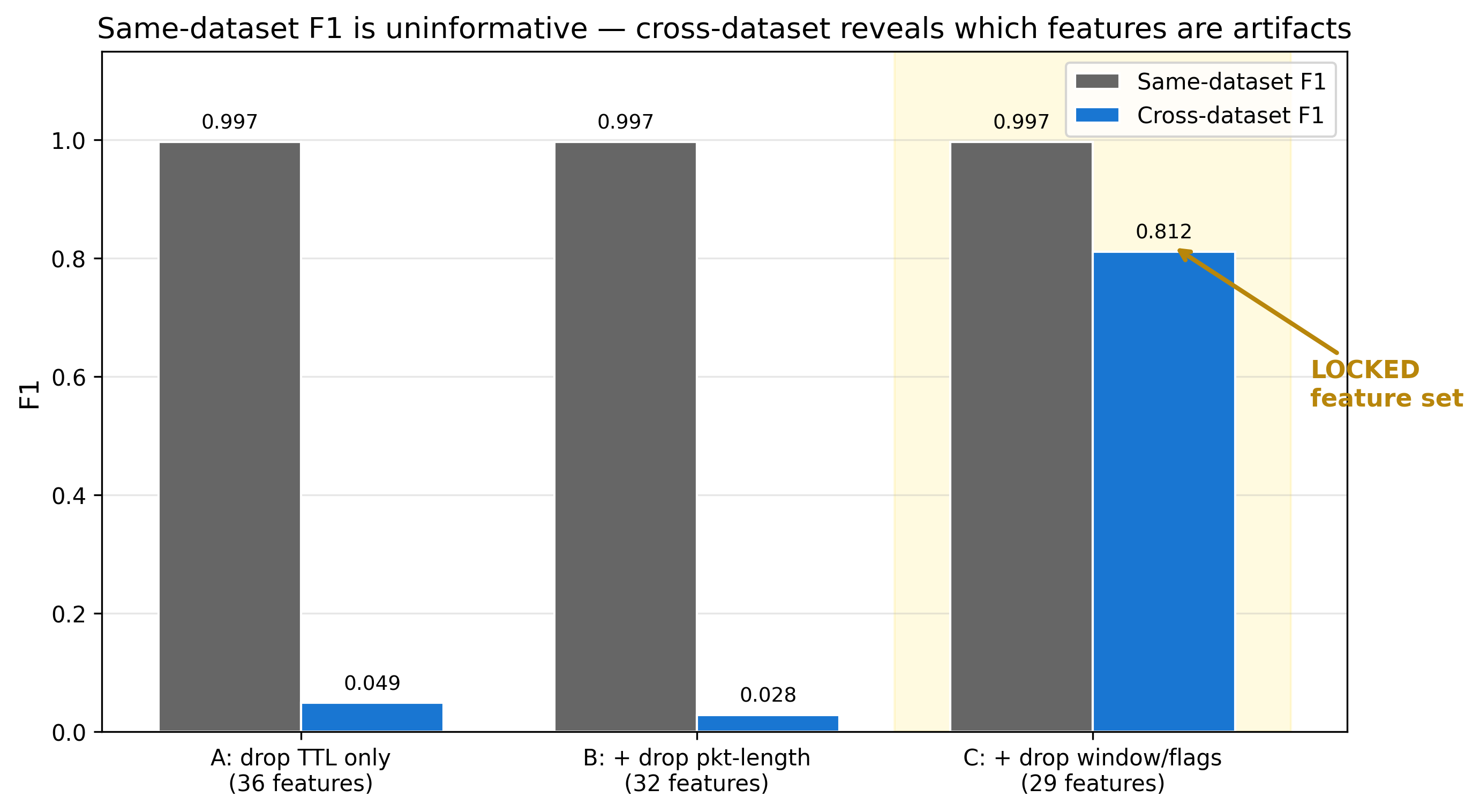}
  \caption{Same-dataset F1 (grey) is flat across all three ablation
  sets while cross-dataset F1 (blue) swings non-monotonically,
  recovering only at Set C. Same-dataset accuracy cannot reveal which
  features leak.}
  \label{fig:ablation}
\end{figure}

Removing leakage features does not improve cross-dataset transfer
monotonically. Set B, which drops more features than Set A, performs
worse on cross-dataset F1 (0.0282 vs.\ 0.0488) and carries a
false-negative rate of 0.9851, missing nearly every attack in the
target domain. Only the aggressive Set C recovers transfer
performance (0.8116 F1, FNR 0.1293), while same-dataset F1 stays
essentially flat across all three sets (0.9966--0.9972). The leakage
signal is therefore not concentrated in the TTL and packet-length
features but distributed across the feature representation. Partial
removal can strip a competing, transferable signal while leaving
enough leakage-adjacent structure intact to mislead the model on the
target domain.

\subsubsection{Transfer is asymmetric under the locked feature set}

\begin{figure}[!tbp]
  \centering
  \includegraphics[width=\linewidth]{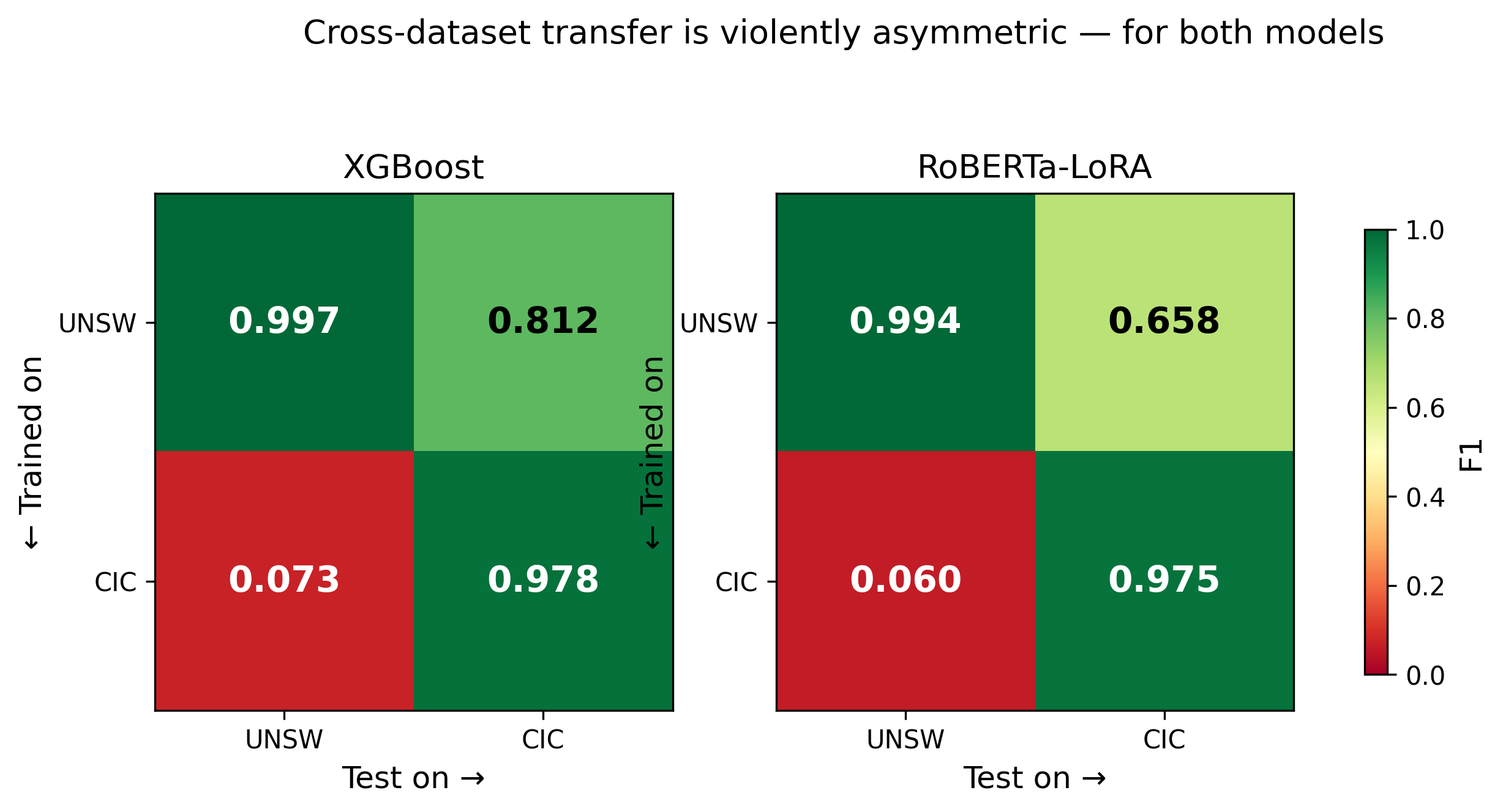}
  \caption{Cross-dataset transfer across both directions, for both
  models. On-diagonal (same-dataset) cells are near-perfect; the
  off-diagonal CIC$\to$UNSW cell collapses for both models, while
  UNSW$\to$CIC remains workable for XGBoost. The asymmetry is
  architecture-independent.}
  \label{fig:transfer-matrix}
\end{figure}

Using Set C, Table~\ref{tab:cross-dataset} reports both transfer
directions for all three models, and Fig.~\ref{fig:transfer-matrix}
shows the full matrix.

\begin{table}[!tbp]
\caption{Cross-dataset F1 under Set C (29 features)}
\label{tab:cross-dataset}
\centering
\footnotesize
\begin{tabular}{lccc}
\toprule
Direction & XGBoost & Random Forest & RoBERTa-LoRA \\
\midrule
UNSW $\to$ CIC & 0.8116 & 0.5662--0.7587 & 0.6582 \\
CIC $\to$ UNSW & 0.0730 & 0.0201--0.0378 & 0.0599 \\
\bottomrule
\end{tabular}
\end{table}

XGBoost beats RoBERTa-LoRA by 15.3 points of F1 in the forward
direction (0.8116 vs.\ 0.6582), and the F1 column understates the
gap: Table~\ref{tab:cd-detail} decomposes it.

\begin{table}[!tbp]
\caption{Cross-dataset UNSW$\to$CIC, decomposed. F1 conceals a near-chance classifier.}
\label{tab:cd-detail}
\centering
\footnotesize
\begin{tabular}{lccc}
\toprule
Metric & XGBoost & RoBERTa-LoRA & Random Forest \\
\midrule
F1                & 0.8116 & 0.6582 & 0.5662--0.7587 \\
Precision         & 0.7601 & 0.5290 & 0.5546--0.6932 \\
Recall            & 0.8707 & 0.8709 & 0.5783--0.8380 \\
FPR               & 0.2748 & \textbf{0.7756} & 0.3710--0.4852 \\
Balanced accuracy & \textbf{0.7979} & \textbf{0.5477} & 0.5569--0.7335 \\
\bottomrule
\end{tabular}
\end{table}

The two models achieve statistically indistinguishable recall on the
target network (0.8707 vs.\ 0.8709). The entire F1 gap comes from
precision, and the FPR row shows why: RoBERTa-LoRA labels 38{,}780 of
50{,}000 benign CIC flows as attacks, a false positive rate of
0.7756. Its balanced accuracy on the target network is 0.5477, barely
above chance, against XGBoost's 0.7979. Read through balanced
accuracy, RoBERTa-LoRA's cross-dataset transfer is close to
non-functional: it retains sensitivity to attacks only by flagging
most normal traffic as well. Random Forest sits between the two on
every metric.

XGBoost shows zero variance across seeds in cross-dataset conditions
(std~=~0.000, Table~\ref{tab:seed-variance}), while Random Forest
varies substantially (std 0.0830, range 0.5662--0.7587
UNSW$\to$CIC). This gap in variance is a direct consequence of how
each model is evaluated. The cross-dataset evaluation uses the full
target dataset for testing with no train/test split randomness, so
XGBoost with fixed hyperparameters produces deterministic predictions
on a fixed 100k-row test set. Random Forest's variance, by contrast,
comes from bagging randomness in tree construction, which XGBoost's
sequential boosting does not have in the same form.

\begin{table}[!tbp]
\caption{Seed variance, cross-dataset UNSW$\to$CIC (seeds 42--45)}
\label{tab:seed-variance}
\centering
\footnotesize
\begin{tabular}{lcc}
\toprule
Model & F1 range & Std \\
\midrule
XGBoost       & 0.8116 -- 0.8116 & 0.0000 \\
Random Forest & 0.5662 -- 0.7587 & 0.0830 \\
\bottomrule
\end{tabular}
\end{table}

The CIC$\to$UNSW direction collapses for every model (0.0201--0.0730
F1). Here balanced accuracy confirms the F1 reading rather than
reinterpreting it: XGBoost reaches 0.5050 and RoBERTa-LoRA 0.5042,
both indistinguishable from chance, with recall of 0.0390 and 0.0316
respectively. Both models fail by the opposite mechanism to the
forward direction, predicting almost nothing as an attack.
Sec.~\ref{sec:discussion} discusses candidate mechanisms for the
asymmetry.

\FloatBarrier

\subsection{Adversarial Evasion}
\label{sec:results-adv}

\begin{figure}[!tbp]
  \centering
  \includegraphics[width=\linewidth]{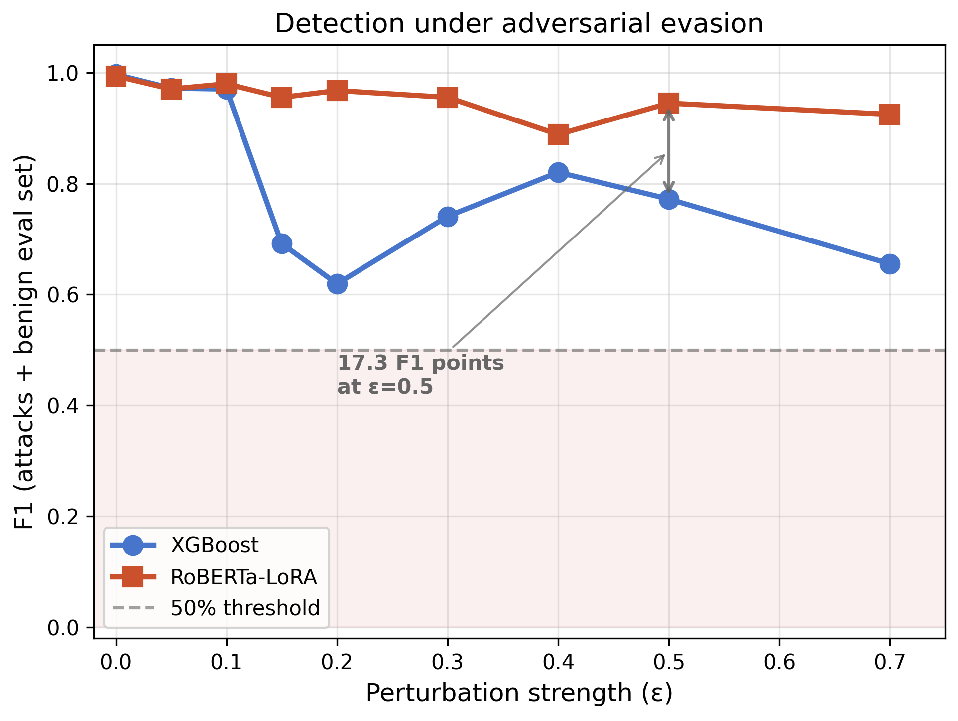}
  \caption{F1 vs.\ perturbation strength $\varepsilon$ on the balanced
  evaluation set (15{,}000 perturbed attack flows plus 15{,}000
  untouched benign flows). The degenerate $\varepsilon{=}1.0$ endpoint
  is excluded.}
  \label{fig:evasion-curve}
\end{figure}

Fig.~\ref{fig:evasion-curve} plots F1 and false positive rate against
perturbation strength $\varepsilon$ for XGBoost and RoBERTa-LoRA
under the feature-space drift-to-benign attack (same-dataset,
UNSW-NB15 only, 15{,}000 perturbed attacks plus 15{,}000 untouched
benign flows). Table~\ref{tab:evasion} reports the full sweep.

\begin{table}[!tbp]
\caption{Adversarial evasion by $\varepsilon$: F1 and FPR on a balanced eval set (15k perturbed attacks + 15k untouched benign)}
\label{tab:evasion}
\centering
\footnotesize
\begin{tabular}{lcccc}
\toprule
& \multicolumn{2}{c}{XGBoost} & \multicolumn{2}{c}{RoBERTa-LoRA} \\
\cmidrule(lr){2-3}\cmidrule(lr){4-5}
$\varepsilon$ & F1 & FPR & F1 & FPR \\
\midrule
0.00 & 0.9966 & 0.0057 & 0.9938 & 0.0081 \\
0.05 & 0.9718 & 0.0057 & 0.9703 & 0.0081 \\
0.10 & 0.9702 & 0.0057 & 0.9797 & 0.0081 \\
0.15 & 0.6917 & 0.0057 & 0.9552 & 0.0081 \\
0.20 & 0.6191 & 0.0057 & 0.9675 & 0.0081 \\
0.30 & 0.7408 & 0.0057 & 0.9553 & 0.0081 \\
0.40 & 0.8202 & 0.0057 & 0.8884 & 0.0081 \\
0.50 & 0.7720 & 0.0057 & 0.9448 & 0.0081 \\
0.70 & 0.6552 & 0.0057 & 0.9247 & 0.0081 \\
\midrule
\multicolumn{5}{l}{\emph{Degenerate endpoint (1 distinct attack row):}} \\
1.00 & 0.0000 & 0.0057 & 0.9959 & 0.0081 \\
\bottomrule
\end{tabular}
\end{table}

The first and most important observation is that neither model trades
false positives for detection. FPR is 0.0057 for XGBoost and 0.0081
for RoBERTa-LoRA at every $\varepsilon$, because benign traffic is
not perturbed. RoBERTa-LoRA's sustained detection therefore reflects
robustness rather than a tendency to label everything an attack. The
same diagnostic, applied to the cross-dataset axis in
Sec.~\ref{sec:results-cd}, returned the opposite verdict for the same
model.

The two models also differ in the magnitude of degradation rather
than its smoothness. Across the valid range ($\varepsilon \le 0.7$)
RoBERTa-LoRA moves within a band of roughly 10 points, from 0.9938 at
$\varepsilon=0$ down to 0.8884 at $\varepsilon=0.40$ and back to
0.9247 at $\varepsilon=0.70$. XGBoost swings across roughly 38
points, from 0.9966 down to 0.6191 at $\varepsilon=0.20$, back up to
0.8202 at $\varepsilon=0.40$, and down again to 0.6552 at
$\varepsilon=0.70$. Both curves are non-monotonic; the contrast is
one of amplitude.

XGBoost's precision never degrades, remaining between 0.9876 and
0.9944 across the entire sweep while recall falls from 0.9989 to
0.4509. RoBERTa-LoRA's precision is likewise stable (0.9900--0.9919).
Where the attack succeeds, it succeeds by pushing attack flows below
the decision threshold, never by inducing false alarms on benign
traffic.

At the representative mid-range strength $\varepsilon=0.5$,
RoBERTa-LoRA leads XGBoost by 17.3 F1 points (0.9448 vs.\ 0.7720),
and the gap reaches 34.8 points at $\varepsilon=0.20$, where XGBoost
is at its weakest. This axis reverses the Sec.~\ref{sec:results-cd}
ranking: RoBERTa-LoRA wins under adversarial evasion, having trailed
XGBoost by 15.3 points under distribution shift. Random Forest is
excluded from this comparison, as the adversarial evasion axis was
evaluated for XGBoost and RoBERTa-LoRA only.

\FloatBarrier

\subsection{Summary Across Axes}
\label{sec:results-summary}

\begin{figure}[!tbp]
  \centering
  \includegraphics[width=\linewidth]{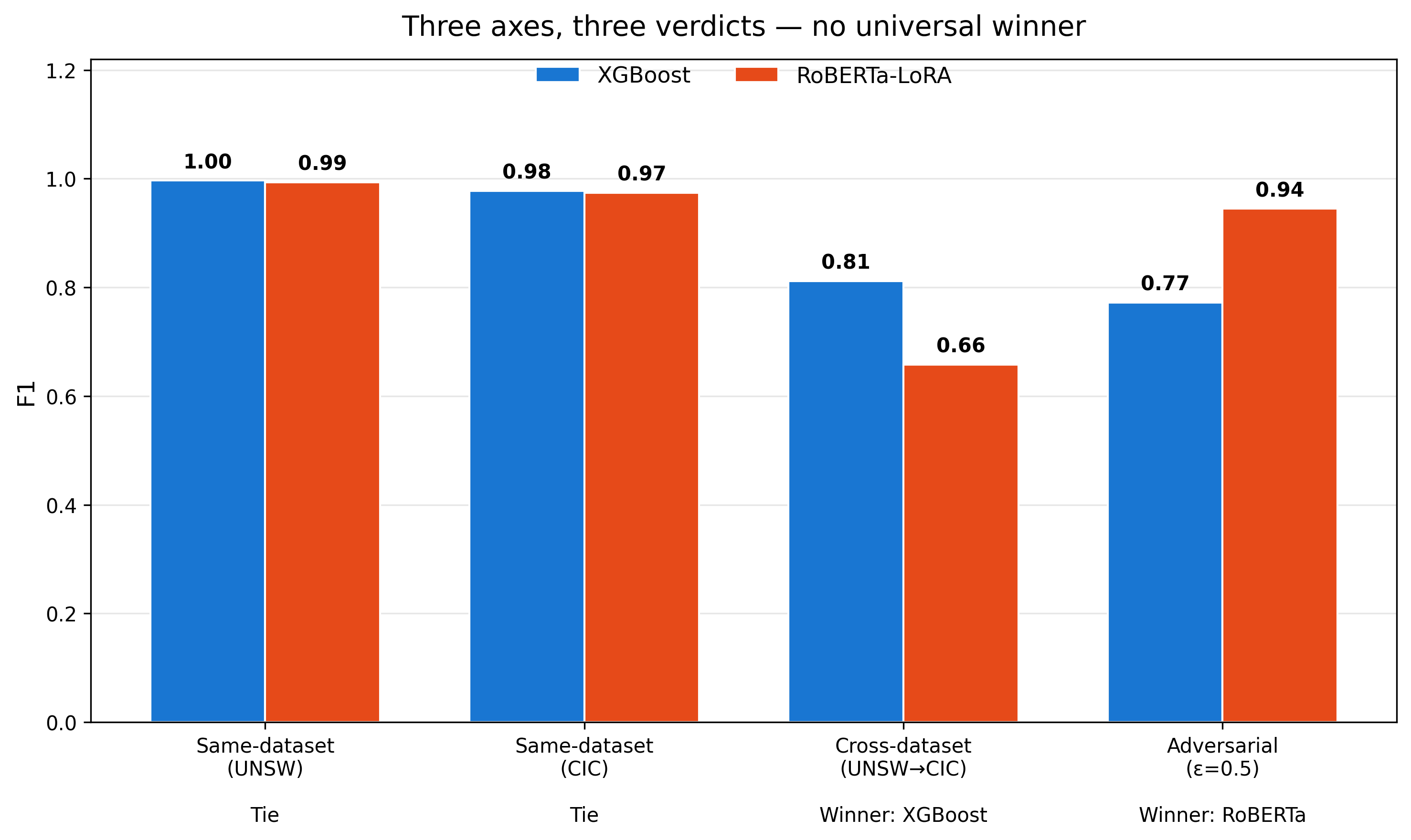}
  \caption{The same two models produce three different verdicts.
  Same-dataset (both networks): statistically tied. Cross-dataset
  (UNSW$\to$CIC): XGBoost wins. Adversarial evasion
  ($\varepsilon=0.5$): RoBERTa-LoRA wins. No single axis predicts the
  others.}
  \label{fig:three-axis}
\end{figure}

Table~\ref{tab:master} collects all three axes for XGBoost and
RoBERTa-LoRA, and Fig.~\ref{fig:three-axis} summarizes the reversal
at a glance. Which model wins depends entirely on which axis is
evaluated. No single benchmark number in that table characterizes
deployment-relevant performance on its own.

\begin{table}[!tbp]
\caption{Master results: F1 across all evaluation axes}
\label{tab:master}
\centering
\footnotesize
\begin{tabular}{lccc}
\toprule
Condition & XGBoost & RoBERTa-LoRA & Gap \\
\midrule
Same-dataset (UNSW)             & 0.9966 & 0.9938 & +0.0028 \\
Same-dataset (CIC)              & 0.9776 & 0.9745 & +0.0031 \\
Cross-dataset (UNSW$\to$CIC)    & 0.8116 & 0.6582 & +0.1534 \\
Cross-dataset (CIC$\to$UNSW)    & 0.0730 & 0.0599 & +0.0131 \\
Adversarial ($\varepsilon=0.2$) & 0.6191 & 0.9675 & $-$0.3484 \\
Adversarial ($\varepsilon=0.5$) & 0.7720 & 0.9448 & $-$0.1728 \\
Adversarial ($\varepsilon=0.7$) & 0.6552 & 0.9247 & $-$0.2695 \\
\bottomrule
\end{tabular}
\end{table}

\FloatBarrier

\section{Discussion}
\label{sec:discussion}

XGBoost and RoBERTa-LoRA cannot be ranked on a single scale.
Same-dataset, they are statistically indistinguishable (within 0.004
F1 on both datasets). Under cross-dataset transfer, XGBoost wins by
roughly 15 F1 points UNSW$\to$CIC and by 25 points of balanced
accuracy, a gap that widens once the metric accounts for false
positives. Under adversarial evasion, RoBERTa-LoRA wins by 17 to 35
F1 points depending on perturbation strength, at comparable and very
low false positive rates. Each axis evaluated alone would support a
confident but incomplete claim about which model is ``better.'' Model
selection for NIDS deployment should therefore be driven by the
anticipated threat model, not by same-dataset benchmark leaderboards.
An operator primarily concerned with deploying across networks unlike
their training data should prefer XGBoost; an operator primarily
concerned with an adaptive adversary who can manipulate flow features
should prefer RoBERTa-LoRA, inference budget permitting
(Sec.~\ref{sec:results-sd}). No evaluation axis in isolation answers
the question a deployer actually has.

A secondary finding cuts across two of the three axes. On the
cross-dataset axis, RoBERTa-LoRA's F1 of 0.6582 looks like degraded
but functional transfer; its balanced accuracy of 0.5477 and FPR of
0.7756 show a classifier close to labelling everything an attack. On
the adversarial axis, RoBERTa-LoRA's sustained F1 could in principle
have had the same explanation. An attacks-only evaluation protocol,
common in this literature, could not have distinguished the two
cases, because precision is unity by construction when the evaluation
set contains no benign traffic. Only the FPR column, measurable
because untouched benign flows are retained at every $\varepsilon$
(Sec.~\ref{sec:adv_evas_attack}), separates robustness (0.0081,
sustained) from the cross-dataset failure mode (0.7756). The same
diagnostic returns opposite verdicts for the same model on two
different axes, which is why adversarial NIDS evaluations should
retain benign traffic in the evaluation set and report FPR alongside
F1 as a matter of protocol.

The three explanations that follow are hypotheses consistent with the
data reported here, not verified mechanisms. Confirming any of them
requires diagnostics the present pipeline does not currently produce:
feature attribution (e.g.\ SHAP) for the ablation account,
per-attack-subtype recall for the transfer asymmetry, and
attention-weight inspection or serialization ablation for the evasion
account. Sec.~\ref{sec:limitations} details what each would require.
They are stated here because they are the leading candidates and
because naming them makes them testable, not because the evidence
settles them.

The non-monotonic ablation result (Table~\ref{tab:ablation}) is
counterintuitive: removing more candidate leakage features (Set B)
produces worse cross-dataset transfer than removing fewer (Set A),
despite same-dataset F1 remaining unchanged across both. One
interpretation is a competing-signal hypothesis. Packet-length
features carry environment-correlated signal, but they also carry
transferable behavioral information, such as coarse differences in
flow size between benign and malicious traffic that hold across
networks. Set A, which removes only TTL, retains this transferable
component alongside the leakage. Set B removes the packet-length
features entirely, discarding the transferable component along with
whatever leakage it carried, and leaves the model with less signal to
generalize on, hence the drop to 0.0282 F1 and a false-negative rate
of 0.9851. Only in Set C, once the broader group of
environment-correlated features is removed together, does the model
apparently learn a representation that no longer depends on any
single discarded feature.

UNSW$\to$CIC transfer is workable for XGBoost (0.8116 F1, 0.7979
balanced accuracy) and marginal for the other two models;
CIC$\to$UNSW collapses to chance for every model. One candidate
explanation, consistent with the two datasets' attack taxonomies
(Sec.~\ref{sec:datasets}), is a difference in attack character.
CSE-CIC-IDS2018's attacks are predominantly volumetric, comprising
DDoS floods, HTTP floods, and brute-force attempts, all producing
flows with anomalously high packet counts, byte counts, or
throughput. A model trained on CIC may learn a decision boundary
close to ``attack = high volume.'' That transfers poorly to
UNSW-NB15, whose attacks (Exploits, Fuzzers, Reconnaissance,
Shellcode, Backdoors) are comparatively quiet and behavioral: a
single malicious probe, or an exploit riding a connection that
otherwise resembles normal traffic. A volumetric decision rule finds
nothing to flag in such traffic, and the observed recall of
0.032--0.039 in that direction is what such a rule would produce. The
reverse direction transfers better under this account because UNSW's
more diverse, lower-volume attack types force a model to learn less
volume-dependent patterns, some of which generalize to CIC's DoS-like
attacks. Computing per-subtype recall in both directions is a direct,
low-cost extension that would resolve this.

XGBoost's F1 swings across roughly 38 points over the valid
perturbation range while RoBERTa-LoRA's moves by about 10. The shape
of XGBoost's curve is consistent with hard, axis-aligned tree splits.
Once an interpolated flow crosses a learned threshold on a small
number of high-importance features, the prediction flips discretely,
producing the sharp fall by $\varepsilon=0.20$ and the partial
recovery at $\varepsilon=0.40$ as different trees' thresholds are
crossed in different directions. That XGBoost's precision holds near
0.99 throughout while recall collapses supports this reading: the
perturbation moves attack flows across the boundary into the benign
region rather than blurring the boundary itself. A complementary
explanation for RoBERTa-LoRA's comparative stability is that flow
serialization (Sec.~\ref{sec:models}) gives it access to structural
patterns in the \texttt{FEATURE=value} text, including feature
ordering, token structure, and the relative magnitude of adjacent
fields, which a purely numeric interpolation perturbs less directly.

\section{Limitations}
\label{sec:limitations}

Every RoBERTa-LoRA result reported in this paper reflects a single
training run (Sec.~\ref{sec:models}), unlike the classical arm, which
is evaluated across four seeds (42--45). RoBERTa figures are reported
as point estimates rather than as a mean with a standard deviation,
and any comparison drawn against XGBoost or Random Forest in
Sec.~\ref{sec:results-summary} should be read with this asymmetry in
mind. A different RoBERTa seed could shift the reported numbers,
particularly in the cross-dataset condition, where the classical
models themselves show meaningfully different variance across seeds
(zero for XGBoost, 0.0830 for Random Forest). Multi-seed LLM
evaluation with proper significance testing is the most direct way to
strengthen the claims in this paper and is left to future work.

RoBERTa checkpoints were selected by F1 on a 4{,}000-flow subset of
the same held-out split subsequently used for reporting
(\texttt{load\_best\_model\_at\_end}, Sec.~\ref{sec:models}). Roughly
13\% of each reported same-dataset evaluation set therefore
participated in model selection, and the same applies to the benign
half of the adversarial evaluation set. The classical arm used no
validation-based model selection, so same-dataset and adversarial
figures for the LLM arm may consequently be marginally optimistic
relative to the classical arm. Cross-dataset figures are unaffected,
as the target network is never seen during training or selection.

The evasion attack in Sec.~\ref{sec:adv_evas_attack} perturbs flow
features directly, subject to domain constraints. This guarantees
that a perturbed flow's \emph{features} are plausible, but it does
not guarantee \emph{problem-space realizability}. An actual attacker
would need to produce raw network traffic whose extracted NetFlow
features match the perturbed values while still achieving the
original objective, such as a successful DDoS or exploit. Pierazzi et
al.~\cite{pierazzi2020problemspace} formalize this distinction, and
their feature-space/problem-space framing is adopted throughout.
Closing the gap typically requires an explicit problem-space attack
construction, as in Venturi et al.'s structural, graph-based evasion
attack against NIDS models~\cite{venturi2024gnnadversarial}, which is
not implemented here. The results in Sec.~\ref{sec:results-adv}
should therefore be read as an upper bound on adversarial robustness,
since some perturbations tested here may not be achievable by an
attacker operating at the packet level. Broader surveys of
adversarial machine learning in NIDS settings similarly emphasize
that feature-space attacks must be evaluated against domain-specific
feasibility constraints to be meaningful~\cite{adversarial_survey_2026}.

The attack drives every attack flow toward one point, the global
benign centroid. This makes the attack cheap to specify and easy to
reproduce, but perturbed flows become progressively less diverse as
$\varepsilon$ grows, degenerating entirely at $\varepsilon=1$
(Sec.~\ref{sec:adv_evas_attack}). A stronger attacker would
interpolate toward a nearby benign flow, or toward a cluster centroid
chosen per attack flow, preserving diversity across the whole range
and plausibly producing a harder evasion set. The reported robustness
gap is specific to this centroid-directed attack and should not be
assumed to generalize to per-flow or cluster-targeted variants.

All findings in this paper are derived from exactly two networks.
This is sufficient to demonstrate that cross-dataset transfer is
directional and asymmetric, but not to establish how that asymmetry
generalizes across networks more broadly. In addition, the pipeline
evaluates only the binary label throughout (Sec.~\ref{sec:datasets}),
so transfer performance cannot be broken down by attack subtype.
Recording attack subtype alongside the binary label, for diagnostic
purposes only rather than as a training target, and computing
per-subtype recall in both transfer directions is a concrete,
low-cost next step that would confirm or overturn the
volumetric-vs-behavioral account.

The LLM arm uses a single architecture, RoBERTa-base with LoRA
adapters (Sec.~\ref{sec:models}). Other encoder architectures, larger
models, and alternative fine-tuning strategies such as full
fine-tuning, QLoRA, or domain-pretrained variants like SecureBERT are
not evaluated. Prior work is mixed on whether this matters. Bui/Boffa
et al.\ found that scaling to a much larger model (Mistral-7B)
provided no benefit over BERT-base ($-0.0\%$ relative to the BERT
baseline), and that domain-specific pretraining helped only
marginally (SecureBERT $+0.7\%$, UniXcoder $+1.1\%$), which suggests
the qualitative findings reported here may generalize across
encoder-style LLMs. This has not been verified for this setting, and
a decoder-only or larger encoder model could behave differently,
particularly on the adversarial evasion axis, where the mechanism
proposed in Sec.~\ref{sec:discussion} is architecture-dependent by
hypothesis.

The distributed nature of the leakage signal uncovered in the
ablation study (Sec.~\ref{sec:feature_abl_study}) means some residual
environment-correlated signal cannot be ruled out in Set C even after
the staged removal. The near-zero movement in same-dataset F1 across
all three ablation sets is consistent with this possibility and was
not independently verified through a feature-attribution method such
as SHAP.

\section{Conclusion}
\label{sec:conclusion}

XGBoost and RoBERTa-LoRA were evaluated across three axes,
same-dataset performance, cross-dataset transfer, and adversarial
evasion, and no universal winner emerged. The two models are tied
same-dataset, XGBoost leads under distribution shift, and
RoBERTa-LoRA leads under adversarial evasion at comparable false
positive rates. Model choice for NIDS deployment should therefore
follow the anticipated threat model rather than a single benchmark
score. A single headline metric also proved insufficient even within
an axis: the same false-positive diagnostic that confirms
RoBERTa-LoRA's adversarial robustness also reveals its cross-dataset
F1 to be concealing a near-chance classifier. Feature-leakage removal
does not improve cross-dataset transfer monotonically, indicating
that the leakage signal is distributed across the feature
representation. Transfer between the two networks studied here is
also strongly directional, though the mechanism behind that
directionality could not be verified with the labels this pipeline
records. Taken together, these findings argue for evaluating NIDS
models, classical or LLM-based, along multiple independent robustness
axes, each assessed with more than one metric.

\bibliographystyle{IEEEtran}
\bibliography{IEEEabrv,refs}

\section*{LLM Usage Statement}

The authors used a large language model (Anthropic Claude) during
the preparation of this work. Its use spanned three areas:
(i) assistance with drafting and editing the manuscript text;
(ii) assistance with implementing, debugging, and refactoring the
experimental pipeline; and (iii) discussion and review of the
experimental design and of the interpretation of results, including
the identification of methodological issues in an earlier version of
the evaluation protocol, specifically the attacks-only adversarial
evaluation set and the degeneracy of the maximal-perturbation
endpoint, which the authors then verified and corrected.

All experiments were designed, executed, and validated by the
authors. All reported numbers were produced by the authors' own
pipeline and independently checked against the recorded outputs. All
cited works were verified by the authors against their published
records. The authors take full responsibility for the content of this
paper.

\end{document}